\documentclass[10pt,twocolumn,letterpaper]{article}

\usepackage{cvpr}              
\usepackage[accsupp]{axessibility} 
\usepackage{multirow}
\usepackage{balance}
\usepackage{cite}
\usepackage{amsmath,amssymb,amsfonts}
\usepackage{graphicx}
\usepackage{multirow}
\usepackage{adjustbox}
\usepackage[table,xcdraw]{xcolor}
\definecolor{mintbg}{rgb}{.63,.79,.95}
\colorlet{lightmintbg}{mintbg!25}

\definecolor{cvprblue}{rgb}{0.21,0.49,0.74}
\usepackage[pagebackref,breaklinks,colorlinks,citecolor=cvprblue]{hyperref}
\usepackage{color}
\def\paperID{4} 
\def\confName{CVPR}
\def\confYear{2024}

\title{Cross-Modal Fusion and Attention Mechanism for Weakly Supervised Video Anomaly Detection}

\author{Ayush Ghadiya\thanks{Equal Contributions} , Purbayan Kar$^*$, Vishal Chudasama$^*$, Pankaj Wasnik\thanks{Corresponding Author.} \\ Media Analysis Group, Sony Research India, Bangalore, India\\
\tt\normalsize\{ayush.ghadiya, purbayan.kar, vishal.chudasama1, pankaj.wasnik\}@sony.com}

\begin{document}
\maketitle

\begin{abstract}
Recently, weakly supervised video anomaly detection (WS-VAD) has emerged as a contemporary research direction to identify anomaly events like violence and nudity in videos using only video-level labels. However, this task has substantial challenges, including addressing imbalanced modality information and consistently distinguishing between normal and abnormal features. In this paper, we address these challenges and propose a multi-modal WS-VAD framework to accurately detect anomalies such as violence and nudity. Within the proposed framework, we introduce a new fusion mechanism known as the Cross-modal Fusion Adapter (CFA), which dynamically selects and enhances highly relevant audio-visual features in relation to the visual modality. Additionally, we introduce a Hyperbolic Lorentzian Graph Attention (HLGAtt) to effectively capture the hierarchical relationships between normal and abnormal representations, thereby enhancing feature separation accuracy. Through extensive experiments, we demonstrate that the proposed model achieves state-of-the-art results on benchmark datasets of violence and nudity detection.
\end{abstract}

\section{Introduction} \label{sec:intro}
In the modern technology era, kids are increasingly turning to online platforms for learning, fun, and connecting with others. However, this easy access also brings up worries about their exposure to harmful and unsuitable content, particularly content with violence and nudity. The potential adverse effects on a child's emotional well-being and psychological development underscores the importance of implementing robust mechanisms to detect violence and nudity.
Detecting such anomalies in a video is a well-known computer vision problem that can also be useful in other real-world applications such as surveillance systems, crime prevention, and content moderation. 
Acquiring annotations for anomalies at the frame level in videos is costly and time-consuming. As a result, 
WS-VAD has emerged as a prominent area of research. WS-VAD focuses on learning abnormal events, such as violence and nudity, solely based on video-level binary labels. In this approach, a video is classified as normal if no anomalous event is detected. In contrast, it is classified as an anomaly if any form of abnormal events, such as violence or nudity, is present.
WS-VAD methods usually employ Multiple Instance Learning (MIL) \cite{MIL} for model training. Here, a regular video is seen as a negative bag with no anomalous segments, while an anomaly video is viewed as positive bag with one or more anomalous segments. The anomaly evaluation function is trained by optimizing the MIL loss to ensure positive bag has a higher anomaly value than negative (normal) bag.

Following MIL, recently, several WS-VAD methods have been proposed based on single-modality (i.e., video-based methods \cite{Real-world_anomaly_detection, wu2021learning, tian2021weakly, li2022self, S3R, tan2024overlooked, karim2024real}) and multi-modality \cite{XDviolence, ICASSP, yu2022modality, HyperVD, UR_DMU, zhang2023exploiting, almarri2024multi}. The multi-modal approaches have shown promising results compared to single-modality-based methods, which jointly learn audio and visual representations to improve performance by leveraging complementary information from different modalities. 
Although multi-modal methods show promising performance, they face two main challenges: 1) unbalanced modality information when combining audio-visual features and 2) inconsistent discrimination between normal and abnormal features. Recently, Peng \emph{et al.} \cite{HyperVD} found that the issue of modality imbalance is mainly due to noise in audio signals from real-world scenarios. To address this, they suggest that auditory information contributes less to anomaly detection than visual cues, leading to lower prioritization of audio features. However, this approach must be corrected when audio data is as crucial as visual data.
To address another issue, i.e., inconsistent discrimination between normal and abnormal features, prior studies have utilized graph representation learning, where each instance is treated as a node in a graph. However, these methods still struggle to distinguish them accurately.

In this study, we propose a new framework to address these challenges. We introduce a novel fusion module called a 
CFA to address the challenge of imbalanced modality information. It dynamically adjusts the influence of each modality by prioritizing the importance of audio features relative to the visual modality. This selective process ensures that only relevant audio features crucial for visual learning are being utilized. By adapting to select the most appropriate features relative to the visual modality, our approach enhances visual feature learning by incorporating relevant audio features. Furthermore, we introduce a hyperbolic graph convolution network-based 
HLGAtt mechanism to maintain consistent discrimination between normal and abnormal features. This mechanism operates in hyperbolic space to capture hierarchical relationships between normal and abnormal representations through spatial and temporal feature learning, which aids in distinguishing normal and abnormal features. 

\begin{figure}[t!]\
    \centering
    \includegraphics[width=\linewidth]{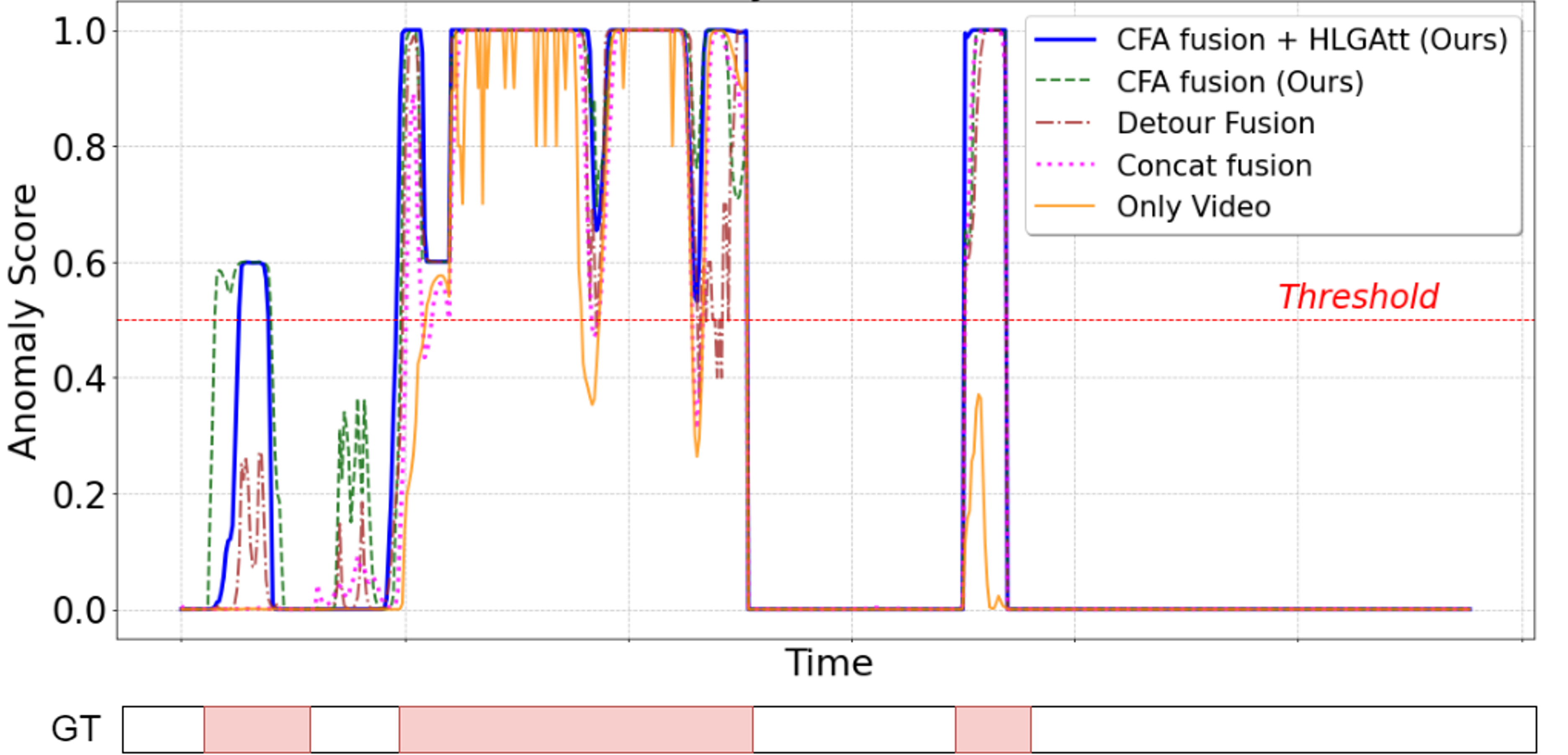}
    \caption{Comparative analysis of our proposed method with prior video-based method as well as audio-video based fusion approaches \cite{XDviolence, HyperVD} on testing videos of XD-Violence dataset.}
    \label{fig:intro}
    \vspace{-1em}
\end{figure}
The proposed model accurately identifies anomaly events and outperforms existing state-of-the-art (SOTA) methods for violence and nudity detection tasks. Figure \ref{fig:intro} shows the anomaly score analysis obtained from a few violent and normal instances of the XD-Violence dataset and compares it with various approaches such as only video-based method \cite{XDviolence}, Concate fusion \cite{XDviolence}, Detour fusion \cite{HyperVD} approaches. Figure \ref{fig:intro} shows that the proposed model accurately identifies anomalies compared to others.
We summarize the contributions of this paper as follows:
\begin{itemize}
    \item We propose a new WS-VAD framework to address the imbalance issue in audio-visual modality information and effectively distinguish abnormal features from normal ones so that anomaly events such as violence and nudity can be detected accurately.
    \item To address the imbalanced modality information issue, we introduce a novel fusion module called 
    CFA, which helps the proposed framework to facilitate multi-modal interaction effectively by dynamically regulating the contribution of each modality.
    \item We introduce a novel attention mechanism called 
    HLGAtt to capture the hierarchical relationships between normal and abnormal representations, thereby enhancing the feature separation.
\end{itemize}

\section{Related Works}\label{sec:literature}
\subsection{Violence Detection Works}
Earlier, few unsupervised learning-based methods \cite{18, sabokrou2018adversarially} have been proposed for violence detection. These methods focus on one-class classification via learning what is normal and spotting anomalies by recognizing deviations from the norm. However, these methods are not well-suited for complex environments and often struggle due to the limited availability of abnormal video data during training. 

Recently, 
WS-VAD methods \cite{Real-world_anomaly_detection, li2022self, XDviolence, yu2022modality, UR_DMU} have been introduced utilizing video-level labels and achieved promising results over unsupervised VAD methods. A few video-based WS-VAD approaches \cite{Real-world_anomaly_detection, wu2021learning, tian2021weakly, li2022self, S3R, tan2024overlooked, karim2024real} have been proposed to enhance the detection accuracy of violence events. However, these approaches overlooked audio information and cross-modality interactions, limiting the effectiveness of violence prediction. To address this issue, Wu \emph{et al.} \cite{XDviolence} introduced a large-scale audio-visual dataset named XD-Violence and established a baseline for audio-visual activities. Following this, many multi-modal approaches \cite{XDviolence, ICASSP, yu2022modality, HyperVD, UR_DMU, zhang2023exploiting, almarri2024multi} have been proposed that outperforms video-based WS-VAD methods. 
Recently, Peng \emph{et al.} \cite{HyperVD} proposed a fusion mechanism for audio-visual data and introduced a hyperbolic graph convolution network-based model to efficiently capture the semantic distinctions via learning the embeddings in hyperbolic space.
Recently, Zhou \emph{et al.} \cite{UR_DMU} proposed a dual memory units module with uncertainty regulation emphasizing learning representations of abnormal and normal data. 
Salem \emph{et al.} \cite{almarri2024multi} introduced a new version of MIL that avoids the disadvantages of ranking loss by using margin loss instead. 

Although these methods present promising results, their effectiveness is hindered by the integration of imbalanced audio-visual features. Moreover, they struggle to consistently differentiate between normal and abnormal features, limiting the detection accuracy. This paper addresses these issues and proposes a new multi-modal framework that detects violent events more accurately. In contrast to recent multi-modal approaches \cite{HyperVD, UR_DMU, zhang2023exploiting}, we propose a new cross-modal fusion with modulation mechanism to learn and fuse audio modality with relative visual features adaptively. Furthermore, we introduce Lorentzian attention-based hyperbolic graph mechanism to learn hierarchical relationships between normal and abnormal features and discriminate them effectively.

\subsection{Nudity Detection Works}
In video-based nudity detection, researchers have devised various methods to tackle the task of identifying explicit content. A common strategy involves detecting skin color in video frames \cite{10Nude,11Nude,12Nude,13Nude}. 
Samal \emph{et al.} \cite{samal2023asyv3} proposed a model that combines attention-enabled pooling with a Swin transformer-based YOLOv3 architecture for obscenity detection in images and videos. 
Jin \emph{et al.} \cite{Deep} employed a weakly supervised multiple instance learning approach for generating a bag of properly sized regions with minimal annotations to tackle the detection of private body parts based on local regions. Wang \emph{et al.} \cite{liyuan2021porn} incorporated an attention-gated mechanism with a deep network, demonstrating its efficacy in performance enhancement. Several studies have proposed deep learning architectures considering local and global context jointly \cite{porn21, Wang2018AdultIC}. Utsav \emph{et al.} \cite{shah2021content} proposed a domain adaptation-based method to filter adult content in streaming video. Tran \emph{et al.} \cite{tran2020additional} proposed an additional training-based approach on pseudo labels using Mask R-CNN for sexual object detection. 

However, above methods focus on image-based approaches or utilize uni-modal approaches; the audio-visual-based approaches have not been extensively explored. This paper seeks to address this gap by employing audio-visual data, aiming to enhance the accuracy of nudity detection in videos.
\section{Methodology} \label{sec:methodology}
\subsection{Problem Statement}
Given a set of $N$ videos, $X = \{X_i\}_{i=1}^N$ and the corresponding ground-truth video-level labels $Y = \{Y_i\}_{i=1}^N \in \{1, 0\}$ where $Y_i=1$ denotes the presence of any abnormal event in the video while $Y_i = 0$ signifies the absence of any abnormal event, we aim to accurately detect abnormal events such as violence and nudity within the videos in a weakly supervised manner. Specifically, each video $X_i$ is initially divided into 16-frame based $T$ non-overlapping multi-modal segments $(M=\{M_i^V, M_i^A\}_{i=1}^T)$, which are processed by a pre-trained CNN network to extract the corresponding visual features $F_V \in \mathbb{R}^{T \times D_V}$ and audio features $F_A \in \mathbb{R}^{T \times D_A}$, where $D_V$ and $D_A$ represents the feature dimensions of video and audio modality. Here, $M_i^V$ and $M_i^A$ denote the video and audio segments, respectively. These extracted visual and audio features are then forwarded into the proposed framework which identifies whether the input video contains any abnormal events or not.

\begin{figure*}[t!]
    \centering
    \includegraphics[width=\linewidth]{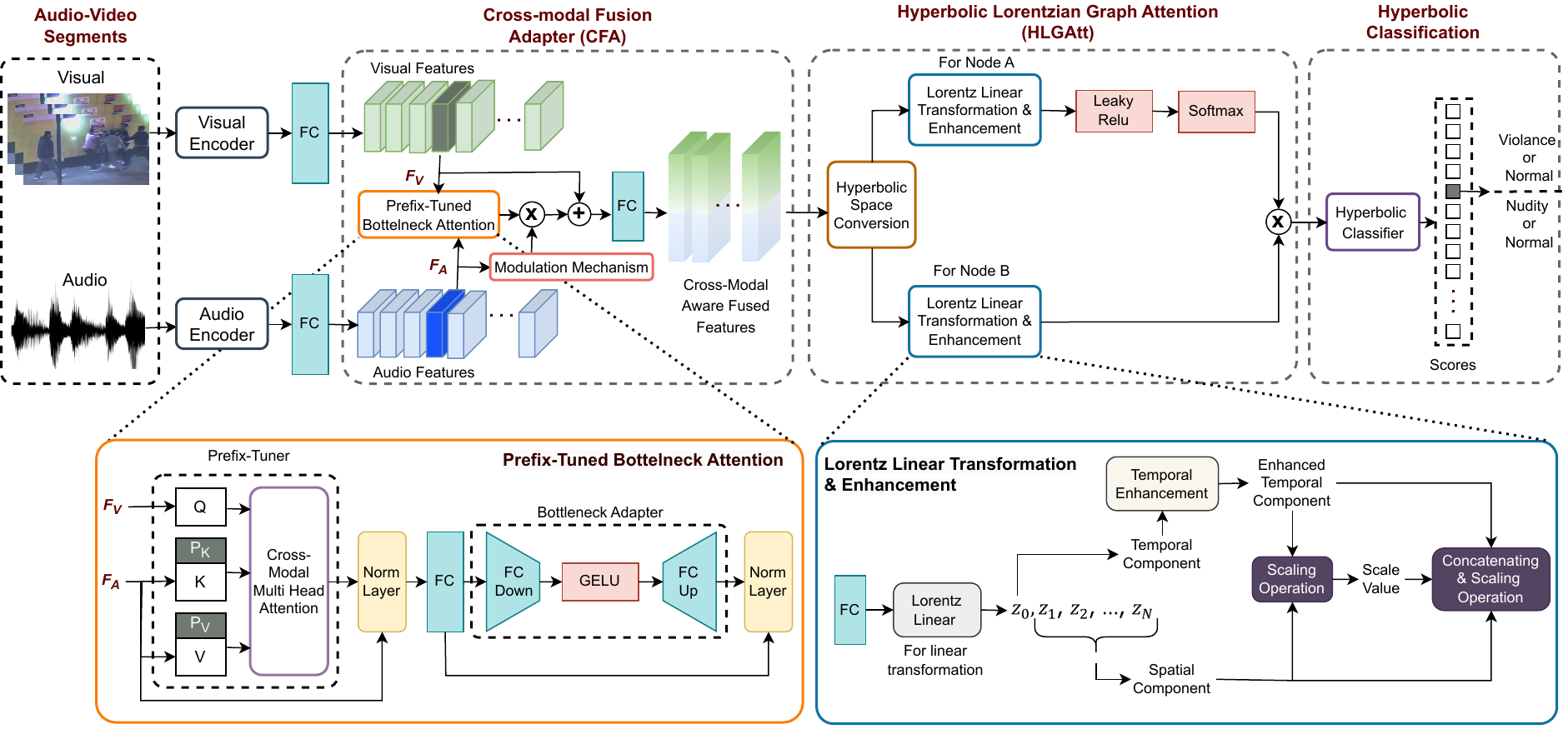}
    \caption{\textbf{Overview of the proposed framework.} It takes audio and visual features extracted from pre-trained encoder networks as input, which are further fused through the proposed Cross-Modal Fusion Adapter (CFA) module to learn multi-modal interaction effectively, followed by the introduced Hyperbolic Lorentzian Graph Attention (HLGAtt) mechanism to capture hierarchical relationships between visual and audio representations, ensuring consistency in distinguishing normal and abnormal features during training. Finally, the outcome features are passed in a hyperbolic classifier to predict anomaly events for each instance.}
    \label{fig:methodology}
\end{figure*}

To identify abnormal events accurately, we propose a new framework as shown in Figure \ref{fig:methodology} in which we introduce novel cross-modal fusion module and hyperbolic Lorentzian graph attention mechanism. Details of these modules are discussed in subsequent subsections. 

\subsection{Cross-modal Fusion Adapter (CFA)}
The CFA module consists of a prefix-tuned-based bottleneck attention and a modulation mechanism. The prefix-tuned bottleneck attention helps in efficient multi-modal interaction between audio and visual modalities. The modulation mechanism dynamically regulates the contribution of each modality during the fusion process, taking into account the importance of the audio features to the visual modality.

\noindent \textbf{Prefix-Tuning bottleneck attention mechanism:} This mechanism incorporates prior knowledge into the feature transformation process by combining the learned representations with initialized parameters through the prefix-tuning operation. 
To do this, the process involves concatenating the keys $K$ and values $V$ obtained from audio features $F_A$ with prefixes $P_k$ \& $P_v$, resulting in prefix-tuned keys $K_p$ and values $V_p$, respectively.
The parameters $P_k$ \& $P_v$ are initialized as zero matrices with dimensions of $\mathbb{R}^{B\times D_A\times D_p}$, where $B$, $D_A$ \& $D_p$ represent the batch size, audio feature dimension, and prefix dimension, respectively.

These prefix-tuned keys $K_p$ and values $V_p$ along with the query $Q$, i.e., visual features $F_V$, are then passed on to the cross-modal multi-head attention module \cite{cross-model}. This module enables the interaction between the prefix-tuned features of the audio and visual modalities, allowing them to selectively and contextually focus on each modality's relevant information. In this process, the attention scores are computed based on queries, prefixed tuned keys and values. The mathematical formulation of the cross-modal multi-head attention module function (i.e., $f_{CMA}$) can be formulated as
\begin{equation}
    F_{Att} = f_{CMA}(Q, K_p, V_p) = Softmax\bigg(\frac{{Q\cdot K_p^T}}{\sqrt{D_{K_P}}} \bigg)\times V_p,
\end{equation}
where, $D_{K_P}$ represents the dimensionality of the key vectors ($K_p$). The attention features $F_{Att}$ are subsequently passed to the bottleneck adapter module. In this stage, the bottleneck adapter ensures smooth interaction between modalities while preserving modality-specific characteristics. It comprises down-scaled fully connected layers (i.e., $f_{down}$) followed by Gaussian Error Linear Unit (GELU) activation (i.e., $f_{GELU}$) and up-scaled fully connected layers (i.e., $f_{up}$). This can be expressed mathematically as

\begin{equation}
    \hat{F}_{Att} = f_{up}(f_{GELU}(f_{down}(F_{Att}))),
\end{equation}
\noindent Here, the GELU activation function introduces non-linearity, allowing intricate feature transformations. This careful design ensures that the adapter module effectively adjusts input features to the shared bottleneck representation, promoting context-aware fusion.

\noindent \textbf{Modulation Mechanism:} 
In the proposed CFA module, we introduce modulation factors that dynamically adjust the impact of individual modalities by considering the importance of their audio features relative to the visual modality. This mechanism is facilitated by a learnable modulation function that operates on audio features $F_A$ to select relevant audio features that are important to visual modality. The resulting modulated features $F_{Mod}$ are defined as
\begin{equation}
    F_{Mod} = f_{MF}(F_A) = \sigma(W_{mod} \cdot F_A).
\end{equation}
\noindent Here, $\sigma$ represents the sigmoid activation, while $W_{mod}$ stands for the weights associated with the modulation function. The sigmoid activation function ensures that modulation factors range between $0$ and $1$, thereby regulating the degree of modulation applied to the fused representation.

Next, the fusion and refinement process is used, where it first fuses the modulated features with the output of the prefix-tuning bottleneck attention and then refines the fused representation through a fully connected layer. This operation can be expressed mathematically as
\begin{equation}\label{eq: Fusion}
    F_{Fused} = f_{FC}(F_V + (\hat{F}_{Att} \times F_{Mod})).
\end{equation}
\noindent The modulation mechanism $f_{MF}()$ modifies the output of the prefix-tuning bottleneck attention based on the significance of their audio features to the visual modality. Through the fusion and refinement process, the final fused representation is carefully crafted to capture the most relevant information from both modalities, simultaneously reducing noise and preserving the modality-specific characteristics.

\subsection{Hyperbolic Lorentzian Graph Attention (HLGAtt)  Mechanism}
In the proposed framework, we introduce a hyperbolic graph convolution network based on a new attention mechanism called 
HLGAtt. The proposed HLGAtt uses a hyperbolic Lorentz graph attention mechanism that learns layer-wise curvature parameters to capture the hierarchical structure of the input graph, thereby enhancing the hierarchical relationship between normal and abnormal representations compared to existing graph-based \cite{XDviolence, HyperVD} or transformer-based \cite{UR_DMU} approaches. 
It consists of a hyperbolic space conversion operation, a Lorentz linear transformation \& enhancement module process on parallel nodes, and a fusing operation.

Initially, we convert the fused audio-visual features $F_{Fused}$ into the hyperbolic space using an exponential function. As a result, we obtain the converted fused features maps $F_{H} \in \mathbb{R}^{T\times 2D_H}$, wherein $T$ denotes the number of segments and $D_H$ represents the hyperbolic dimension.

Recently, Zhang \emph{et al.} \cite{zhang2021hyperbolic} proposed a hyperbolic graph attention mechanism that utilized a parallel branch process to learn different features and patterns in respective branches for prediction tasks.
Inspired by this \cite{zhang2021hyperbolic}, we process the converted hyperbolic feature maps on two parallel branches, i.e., node A and node B, to learn specific patterns from the input feature maps. Separating the branches ensures that features with similar characteristics are directed to their respective nodes. This allows each branch to learn the unique properties of normal and abnormal features, enabling more precise discrimination between them. 


The converted hyperbolic feature maps are passed through the Lorentzian linear transformation \& enhancement module in each node. Here, we employ the Lorentzian linear transformation \cite{chen2021fully,HyperVD} for feature transformation and its transformed temporal and spatial features are further enhanced using the proposed enhancement mechanism. In Lorentzian linear transformation, we first establish the adjacency matrix $A \in \mathbb{R}^{T \times T}$ to capture hyperbolic feature similarities. Here, each entry $A_{ij}$ can be calculated as 
\begin{equation}    
    \begin{aligned}
        A_{ij} &= f_{sim}(F_{H, i}, F_{H, j}) \\
           &= Softmax(\exp(-d_L(F_{H, i}, F_{H, j})),   
    \end{aligned}
\end{equation}
where, $f_{sim}$ represents the hyperbolic feature similarity measure, which evaluates how closely snippets $i$ and $j$ resemble each other based on their Lorentzian intrinsic distance $d_L$. The exponential and Softmax functions are employed to maintain non-negativity and restrict the values of $A$ within the range of $[0, 1]$.

Next, we incorporate a hyperbolic Lorentz linear (i.e., $f_{HL}()$), followed by neighborhood hyperbolic aggregation operation \cite{qu2023hyperbolic} for feature transformation. These transformed hyperbolic features of the $i^{th}$ snippet at the layer $l$ (i.e., $z_{i}^l$) can be expressed as
\begin{equation}
   z_i^l = F_{H, i}^l=\frac{\sum_{j=1}^T A_{i j} f_{HL}\left(F^{l-1}_{H, i}\right)}{\sqrt{-\eta}\left|\left\|\sum_{k=1}^T A_{i k} f_{HL}\left(F^{l-1}_{H, i}\right)\right\|_{\mathcal{L}}\right|},
\end{equation}
where, $\eta$ indicates the negative curvature constant. 

To enhance these transformed features $z$ further, they are processed based on temporal and spatial information. The initial component of the input vector $z[0]$ signifies the temporal aspect within hyperbolic space \cite{chen2021fully}. This component is processed via a sigmoid activation function followed by exponential scaling and shifting operations. Through this procedure, temporal features (i.e., $T_{nodeA}$ and $T_{nodeB}$) are computed for both node A and node B as 
\begin{equation}    
    \begin{aligned}
        T_{nodeA} &= \sigma(z_{nodeA}[0]) \times e^{{\gamma}} + 1.1 \\
        T_{nodeB} &= \sigma(z_{nodeB}[0]) \times e^{{\gamma}} + 1.1    
    \end{aligned}
\end{equation}
where, $\gamma$ is a trainable parameter. The remaining elements of input vector $z$ can be considered as the spatial features \cite{chen2021fully} for node A and node B (i.e., $S_{nodeA}$ and $S_{nodeB}$). Mathematically, they can be formulated as
\begin{equation}    
    \begin{aligned}
        S_{nodeA} &= [z_{nodeA}[1],z_{nodeA}[2], ... , z_{nodeA}[n]] \\
        S_{nodeB} &= [z_{nodeB}[1],z_{nodeB}[2], ... , z_{nodeB}[n]]    
    \end{aligned}
\end{equation}
These features encapsulate the intricate spatial features in hyperbolic space, which are critical for capturing the hierarchical structure and relationships within the graph.

To ensure the alignment of spatial components with the hyperbolic model, a scaling factor, referred to as $\Upsilon$ is computed. This factor takes into account the temporal and spatial complexities of each node. It ensures that the spatial components are appropriately scaled to fit within the hyperbolic space.
\begin{equation}    
    \begin{aligned}
        \Upsilon_{nodeA} = \frac{{T_{nodeA}}^2 - 1}{{\sum_{i=1}^{n} (S_{nodeA}[i])^2} + \epsilon}  \\
        \Upsilon_{nodeB} = \frac{{T_{nodeB}}^2 - 1}{{\sum_{i=1}^{n} (S_{nodeB}[i])^2} + \epsilon}   
    \end{aligned}
\end{equation}

The temporal and scaled spatial components are concatenate, resulting in enhanced feature vectors (i.e., $\hat{F}^{nodeA}_{H}$ and $\hat{F}^{nodeB}_{H}$). Mathematically, this process can be expressed as
\begin{equation}    
    \begin{aligned}
        \hat{F}^{nodeA}_{H} &= Concat\Big[T_{nodeA}, S_{nodeA} \times \sqrt{\Upsilon_{nodeA}}\Big]  \\
        \hat{F}^{nodeB}_{H} &= Concat\Big[T_{nodeB}, S_{nodeB} \times \sqrt{\Upsilon_{nodeB}}\Big]   
    \end{aligned}
\end{equation}
The enhanced feature maps in the node A branch passed through Leaky-ReLU activation and softmax normalization operations to introduce non-linearity and ensure standardization across the enhanced feature maps. This ensures that distinct patterns, representing normal and abnormal data, are learned at each node. By doing so, the model is encouraged to learn different sets of features from those processed by other node (i.e., node B). Finally, the enhanced feature maps from node A and node B are processed via matrix multiplication to compute attention, followed by a ReLU activation to generate the output feature maps. This outcome of the proposed HLGAtt module can be formulated as
\begin{equation}
    F_{H}^{final} = f_{ReLU}(\hat{F}^{nodeA}_{H} \cdot \hat{F}^{nodeB}_{H}).
\end{equation}

\subsection{Hyperbolic Classifier \& Learning Objective}
Following \cite{HyperVD}, we also utilize the hyperbolic classifier, which takes the output of the HLGAtt module as input and predicts the confidence scores for normal and abnormal events. The final score $Score$ can be represented as
\begin{equation}
      Score = f_{Hyp-cls}(F_{H}^{final})
\end{equation}
In order to train the proposed model end-to-end, we employ the MIL-based learning objective adopted in \cite{Real-world_anomaly_detection, XDviolence, wtac, HyperVD}, which calculates the mean value of the top $k-$max predictive scores within a video. The high-scoring positive predictions indicate the presence of abnormal events, while the $k-$max negative scores usually represent hard samples. This learning objective function can be formulated as 
\begin{equation}\label{eq: loss}
    L_{MIL} = \frac{1}{N}\sum_{i=1}^{N} -Y_i\cdot \log(\overline{Score}).
\end{equation}
Here, $\overline{Score}$ indicates the average of the $k-$max scores in the video, and $Y_i$ represents the binary video-level label.

\section{Experiments and Results}
\label{sec:result_analysis}
\subsection{Implementation Details}
The proposed model is trained/tested on benchmark XD-Violence dataset \cite{XDviolence} for violence detection task, on NPDI pornography dataset \cite{NPDI, NPDI1} for nudity detection task. The details of these datasets are mentioned below:
\begin{itemize}
    \item \textbf{XD-Violence for violence detection:} The XD-Violence dataset \cite{XDviolence} is a diverse compilation of 4754 raw videos (equivalent to 217 hours) gathered from real-world sources, including movies, web videos, sports broadcasts, security cameras, and CCTVs. It consists of six types of violent events, such as abuse, auto crashes, and shootings, with corresponding video-level annotations. The testing set comprises 300 normal and 500 violent videos, while the training set includes 2049 normal and 1905 violent videos, all labeled at the video level.
    \item \textbf{NPDI for Nudity Detection :} The NPDI Pornography benchmark dataset \cite{NPDI, NPDI1} comprises around 80 hours of video content extracted from 400 movies. These contents are classified as pornographic or non-pornographic, with an equivalent amount of videos in each category. Within the non-pornographic section, there are 200 videos labeled as either ``easy" or ``difficult". The ``easy" videos were randomly selected, while the ``difficult" ones were obtained through textual search queries such as ``beach," ``wrestling" and ``swimming". Although the ``difficult" videos may contain body skin, they do not include explicit nudity or pornographic content.
\end{itemize}
\textbf{Training / Evaluation Details:} The proposed model is trained on datasets mentioned above using the multi-instance learning-based loss function (i.e., Eq. \ref{eq: loss}) with a batch size of 128. During the training process, we adopt the Adam optimizer with a learning rate of $5\times 10^{-4}$ varied using a cosine annealing scheduler and trained for 50 epochs. 
For fair comparison with existing SOTA methods, the proposed framework also employs a pre-trained I3D model \cite{i3d} to extract the visual features ($F_V$), while the VGGish network \cite{vgg2} is utilized to extract the audio features ($F_A$). In the proposed framework, we use the LeakyReLU activation function with a negative slope of -2.
In the Prefix-Tuner of the CFA module, we empirically chose the prefix dimension as 64. The bottleneck adapter has a size of 256 and utilizes the GELU activation function with a dropout rate of 0.1. The constant representing negative curvature ($\eta$) is set to -1 during training. 

For comparison on violence detection task, we choose unsupervised methods (i.e., SVM baseline, and Hasan \emph{et al.} \cite{hasan2016learning}), video modality-based weakly supervised methods \cite{Real-world_anomaly_detection, XDviolence, tian2021weakly, li2022self, S3R, UR_DMU, Real-world_anomaly_detection, tan2024overlooked, zhang2023exploiting}, and audio-visual modality-based weakly supervised methods \cite{XDviolence, ICASSP, yu2022modality, UR_DMU, almarri2024multi, zhang2023exploiting}). The frame-level average precision (AP) metric is adopted to compare these methods, whereas a higher AP measure means better performance. For the nudity detection task, we compare the proposed method with existing methods \cite{Deep, tran2020additional, samal2023asyv3, HyperVD, shah2021content, yahoo}.
However, these methods have utilized uni-modal approaches in their network. 
Additionally, we re-train the recent multi-modal SOTA method called HyperVD \cite{HyperVD} on the NPDI dataset. For comparison, we use the standard evaluation metrics, i.e., AP, accuracy, precision, and recall, where higher measures of these evaluation metrics indicate superior performance.

All the experiments were implemented using PyTorch and the network was trained on a 40GB NVIDIA A100 GPU with batch size of 128. 

\subsection{Result Analysis on Violence Detection task}
Table \ref{tab:Violence} compares state-of-the-art methods on the XD-Violence testing dataset in terms of AP metric. Notably, our proposed method outperforms both video modality-based and audio-video modality-based methods. It achieves an AP score of 86.34\%, which is 0.67\% higher than the previous best-performing method HyperVD \cite{HyperVD}. Compared to video-modality-based methods, our proposed approach shows a 4.24\% 
\begin{table}[t]
\centering
\caption{Comparison against SOTA methods on XD-Violence Dataset for violence detection. Best result is \textbf{bolded} and second best result is \underline{underlined}.}
\label{tab:Violence}
\vspace{-0.5em}
\begin{adjustbox}{max width=0.975\linewidth}
\begin{tabular}{|l|l|c|c|}
 \hline
\rule{-2pt}{5pt} Method        & Publication          & Modality       & AP (\%)        \\ \hline
\rowcolor{lightmintbg}\multicolumn{4}{|l|}{Unsupervised learning based methods} \\ \hline
\rule{-2pt}{5pt} SVM baseline  & NIPS'99         & Video             & 50.78          \\
  \rule{-2pt}{5pt} Hasan \emph{et al.} \cite{hasan2016learning}   &  CVPR'16      & Video             & 30.77          \\\hline
\rowcolor{lightmintbg} \multicolumn{4}{|l|}{Weakly supervised learning based methods} \\ \hline
\rule{-2pt}{5pt} Sultani \emph{et al.} \cite{Real-world_anomaly_detection} & CVPR'18 & Video              & 75.68          \\
\rule{-2pt}{5pt}    Wu \emph{et al.} \cite{XDviolence}  &  ECCV'20 & Audio + Video          & 78.66          \\
\rule{-2pt}{5pt}     Wu \emph{et al.} \cite{wu2021learning}  &  ICIP'21    & Video              & 75.90          \\
 \rule{-2pt}{5pt}     Pang \emph{et al.} \cite{ICASSP} & ICASSP'21   & Audio + Video          & 81.69          \\
\rule{-2pt}{5pt}     RTFM \cite{tian2021weakly}   & ICCV'21      & Video              & 77.81          \\
\rule{-2pt}{5pt}   MSL \emph{et al.} \cite{li2022self} &  AAAI'22   & Video              & 78.28          \\
\rule{-2pt}{5pt}   S3R    \cite{S3R}  &  ECCV'22      & Video              & 80.26          \\
\rule{-2pt}{5pt}       MACIL-SD \cite{yu2022modality}   &  ACM'22   & Audio + Video          & 83.40          \\
\rule{-2pt}{5pt} HyperVD \cite{HyperVD}     & arXiv'23  & Audio + Video          & \underline{85.67}          \\ 
 \rule{-2pt}{5pt}  UR-DMU  \cite{UR_DMU}   & AAAI'23  & Audio + Video           & 81.77          \\
\rule{-2pt}{5pt} Zhang \emph{et al.} \cite{zhang2023exploiting}  & CVPR'23 & Audio + Video          & 81.43          \\
\rule{-2pt}{5pt}      Salem \emph{et al.} \cite{almarri2024multi} &    WACVW'24    & Audio + Video             & 71.40         \\
\rule{-2pt}{5pt}       Tan \emph{et al.} \cite{tan2024overlooked} &    WACVW'24    & Video             & 82.10         \\
\rule{-2pt}{5pt}       REWARD-E2E \cite{karim2024real} &    WACV'24    & video             & 80.30         \\
\rowcolor{lightmintbg}\rule{-2pt}{5pt}   \textbf{Proposed} & \multicolumn{1}{|c|}{---} & \textbf{Audio + Video} & \textbf{86.34} \\ \hline
\end{tabular}
\end{adjustbox}
\end{table}

\begin{figure}[t]
    \centering
    \includegraphics[width=\linewidth]{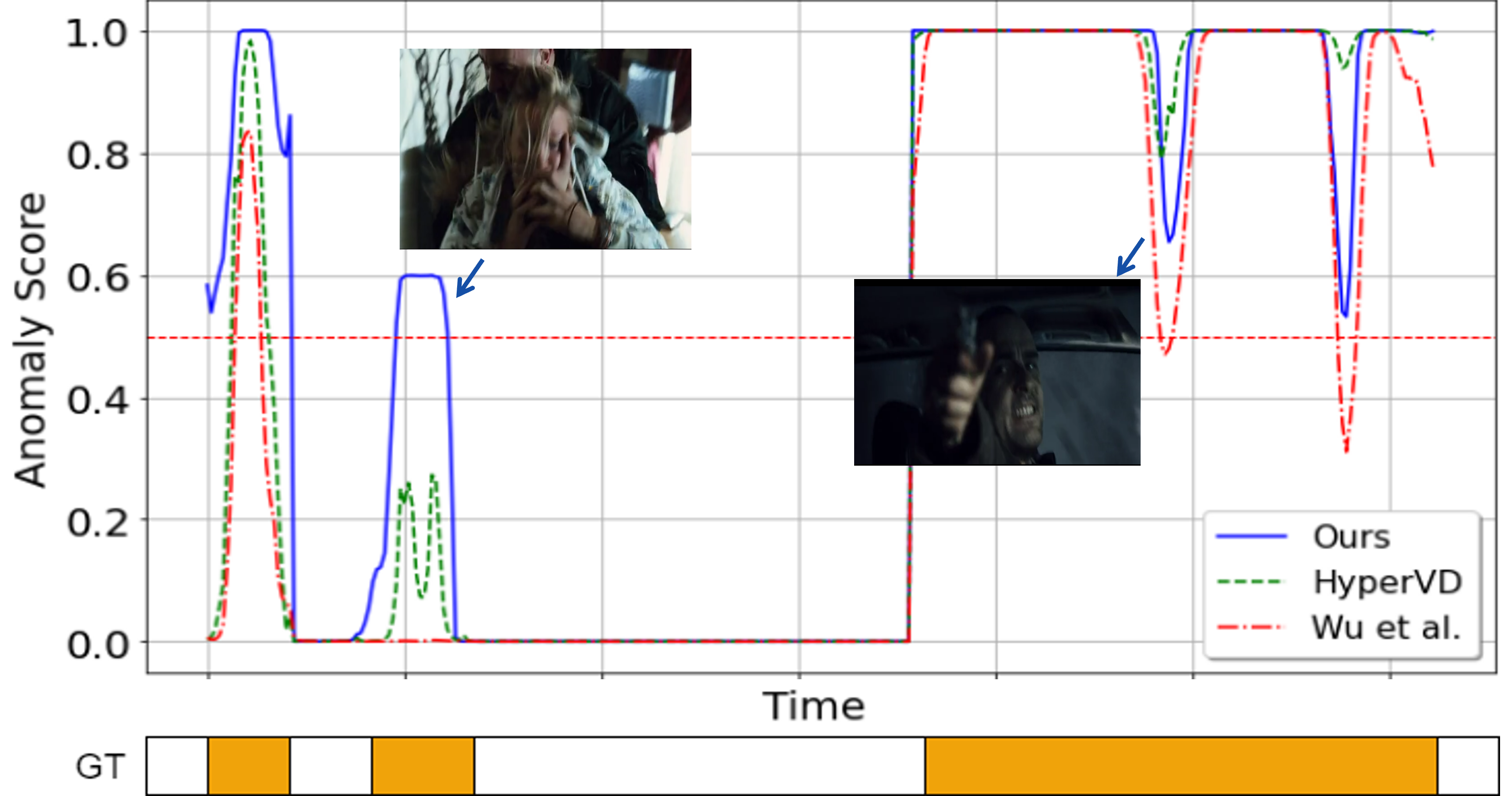}
    \caption{Visual comparison in terms of anomaly score curves on sample video of XD-Violence dataset. Yellow regions are the temporal ground-truths of violent events.}
    \label{fig:violence}
\end{figure}

Figure \ref{fig:violence} displays the visual prediction analysis of our method when compared to existing methods, i.e., HyperVD \cite{HyperVD} and Wu \emph{et al.} \cite{XDviolence}. The comparison is based on the anomaly score obtained from a few videos of the XD-Violence testing dataset \cite{XDviolence}. Here, one can observe that the proposed method not only identifies violent event regions but also yields superior and more precise anomaly scores compared to other methods. 

\begin{figure}[t]
    \centering
    \includegraphics[width=\linewidth]{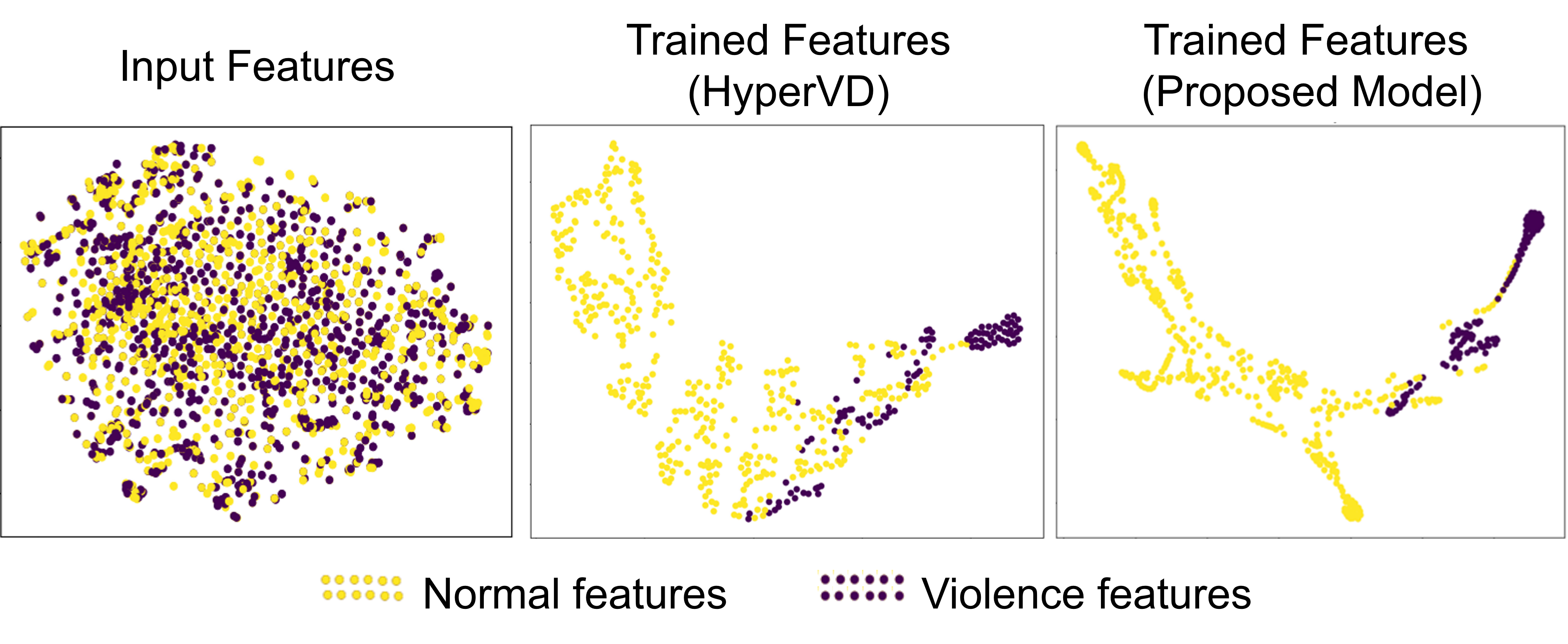}
    \caption{Visual comparison on normal and violence features of the proposed and HyperVD \cite{HyperVD} methods on XD-Violence dataset.}
    \label{fig:tSNE-Violence}
\end{figure}

Additionally, Figure \ref{fig:tSNE-Violence} provides a comparison between the proposed and HyperVD \cite{HyperVD} methods in terms of t-SNE visualization \cite{tSNE} of normal and violent features distributions on the XD-Violence dataset testing videos. One can find that the proposed method effectively clusters the violent and non-violent features and also enlarges the distance between uncorrelated features after the training procedure as compared to the HyperVD \cite{HyperVD} method. 
\subsection{Result Analysis on Nudity Detection task}
This section provides an analysis of the Nudity detection task, comparing it with existing methods \cite{Deep,tran2020additional,samal2023asyv3,HyperVD,shah2021content,yahoo} on the NPDI testing dataset \cite{NPDI, NPDI1}. Table \ref{tab:nudity} shows the comparison in terms of AP, accuracy, precision, and recall. Here, it can be noticed that the proposed model outperforms other methods in all evaluation metrics by a significant margin. To be specific, our model achieves an AP of 99.45\%, an accuracy of 94.12\%, a precision of 95\%, and a recall of 93.75\%. Notably, it demonstrates improvements of at least 1.95\% in AP, 0.42\% in accuracy, 2.2\% in precision, and 3\% in recall compared to other methods.

\begin{table}[t]
\centering
\caption{Comparison against SOTA methods on NPDI Dataset for nudity detection. The best result is \textbf{bolded} and the second best result is \underline{underlined}. Here, * indicates the re-trained method.} \label{tab:nudity}
\vspace{-0.5em}
\begin{adjustbox}{max width=\linewidth}
\begin{tabular}{|l|c|c|c|c|c|}
\hline
\rule{-2pt}{5pt} \textbf{Methods}   & Modality     & \textbf{AP}   & \textbf{Accuracy} & \textbf{Precision} & \textbf{Recall} \\ \hline
\rule{-2pt}{5pt} OpenYahoo \cite{yahoo}    & Video         & 79.0          &---                &---                 &---              \\
\rule{-2pt}{5pt} Deep Region-based CNN \cite{Deep}& Video & 87.8          &---                &---                 &---              \\
\rule{-2pt}{5pt} Deep Part Detector \cite{Deep}  & Video  & 87.0          &---                &---                 &---              \\
\rule{-2pt}{5pt} Deep MIL \cite{Deep}         &  Video    & 86.0          &---                &---                 &---              \\
\rule{-2pt}{5pt} Weighted MIL \cite{Deep}   &   Video    & 97.5          &---                &---                 &---              \\
\rule{-2pt}{5pt} Tran \emph{et al.} \cite{tran2020additional} & Video   & ---             & 90.43            &---                 & ---               \\
\rule{-2pt}{5pt} WD-based adaptation \cite{shah2021content}& Video   & \underline{96.92}         & \underline{93.70}             & ---                  & ---               \\
\rule{-2pt}{5pt} ASYv3 \cite{samal2023asyv3}          & Video       & 89.87         & 89.35             & 89.38              & 89.55         \\  
\rule{-2pt}{5pt} HyperVD*   \cite{HyperVD}         & Audio + Video      & 96.45         & 92.19             & \underline{92.80}              & \underline{90.75}       \\ 
\rowcolor{lightmintbg} \rule{-2pt}{5pt} \textbf{Proposed} & \multicolumn{1}{|c|}{Audio + Video} & \textbf{99.45} & \textbf{94 .12}   & \textbf{95.00}  & \textbf{93.75}  \\ \hline \hline
\end{tabular}
\end{adjustbox}
\end{table}
\begin{figure}[t]
    \centering
    \includegraphics[width=\linewidth]{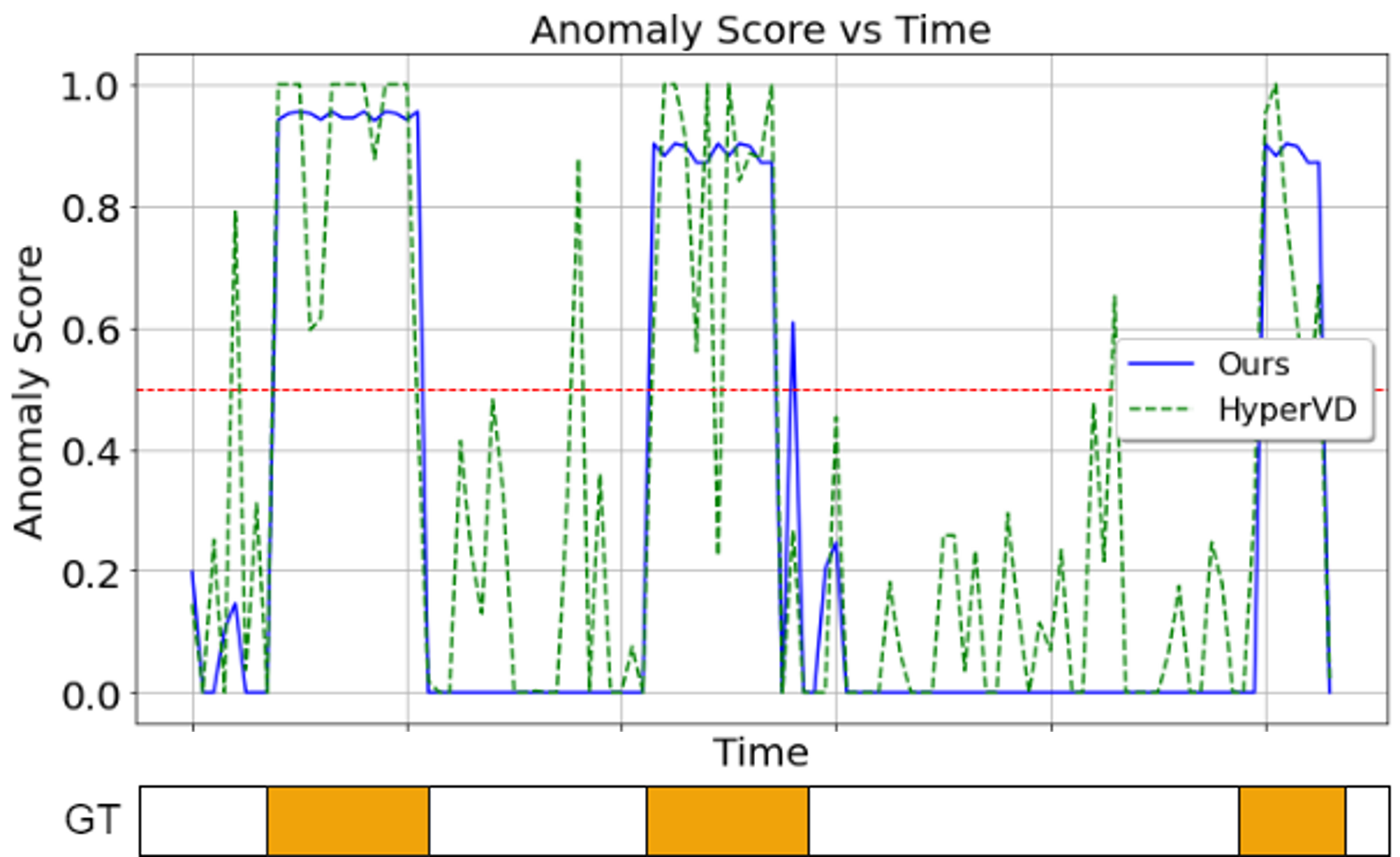}
    \caption{Visual comparison between proposed model and HyperVD \cite{HyperVD} in terms of Anomaly Score vs Time. Yellow regions are the temporal ground-truths of nudity events.}
    \label{fig:nudity}
\end{figure}
\begin{figure}[t]
    \centering
    \includegraphics[width=\linewidth]{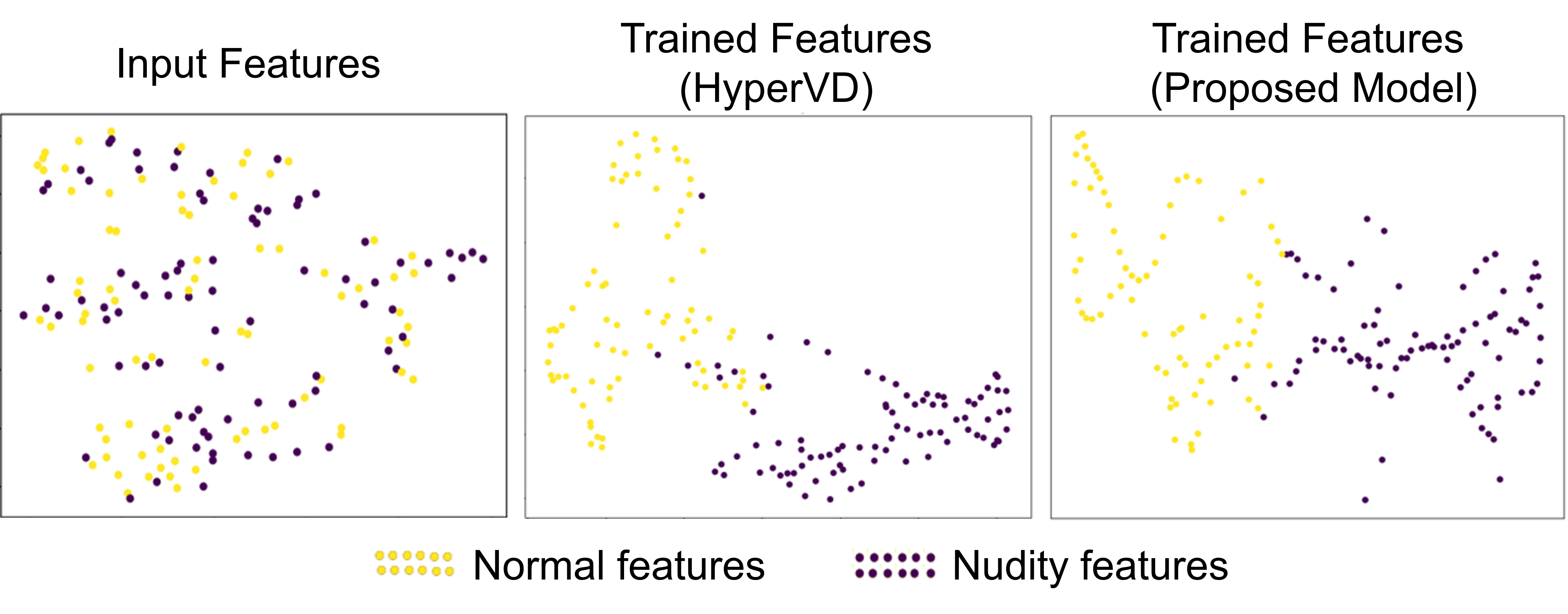}
    \caption{Visual comparison on normal and nudity event features of the proposed and HyperVD \cite{HyperVD} methods on NPDI dataset.}
    \label{fig:tSNE-Nudity}
\end{figure}
Additionally, Figure \ref{fig:nudity} demonstrates the visualization of the anomaly score obtained from the proposed and HyperVD \cite{HyperVD} methods on the NPDI dataset.
The visualization demonstrates that the proposed method generates minimal predictions for regular segments in normal footage while effectively handling extreme situations within nudity content. This analysis also proves that the proposed method not only accurately identifies specific regions but also provides more precise anomaly scores compared to anomaly predictions of the HyperVD method \cite{HyperVD}. 

We also provide t-SNE \cite{tSNE} based visual comparison of the proposed and HyperVD \cite{HyperVD} methods in Figure \ref{fig:tSNE-Nudity}. Here, we compare the proposed and re-trained HyperVD \cite{HyperVD} methods with corresponding normal and violent feature distributions obtained from the NPDI dataset testing videos. It can be clearly observed that the proposed method performs better in clustering the nudity and normal features as compared to the HyperVD \cite{HyperVD} method. 

\subsection{Ablation Analysis}
\begin{figure}
    \centering
    \includegraphics[width=\linewidth]{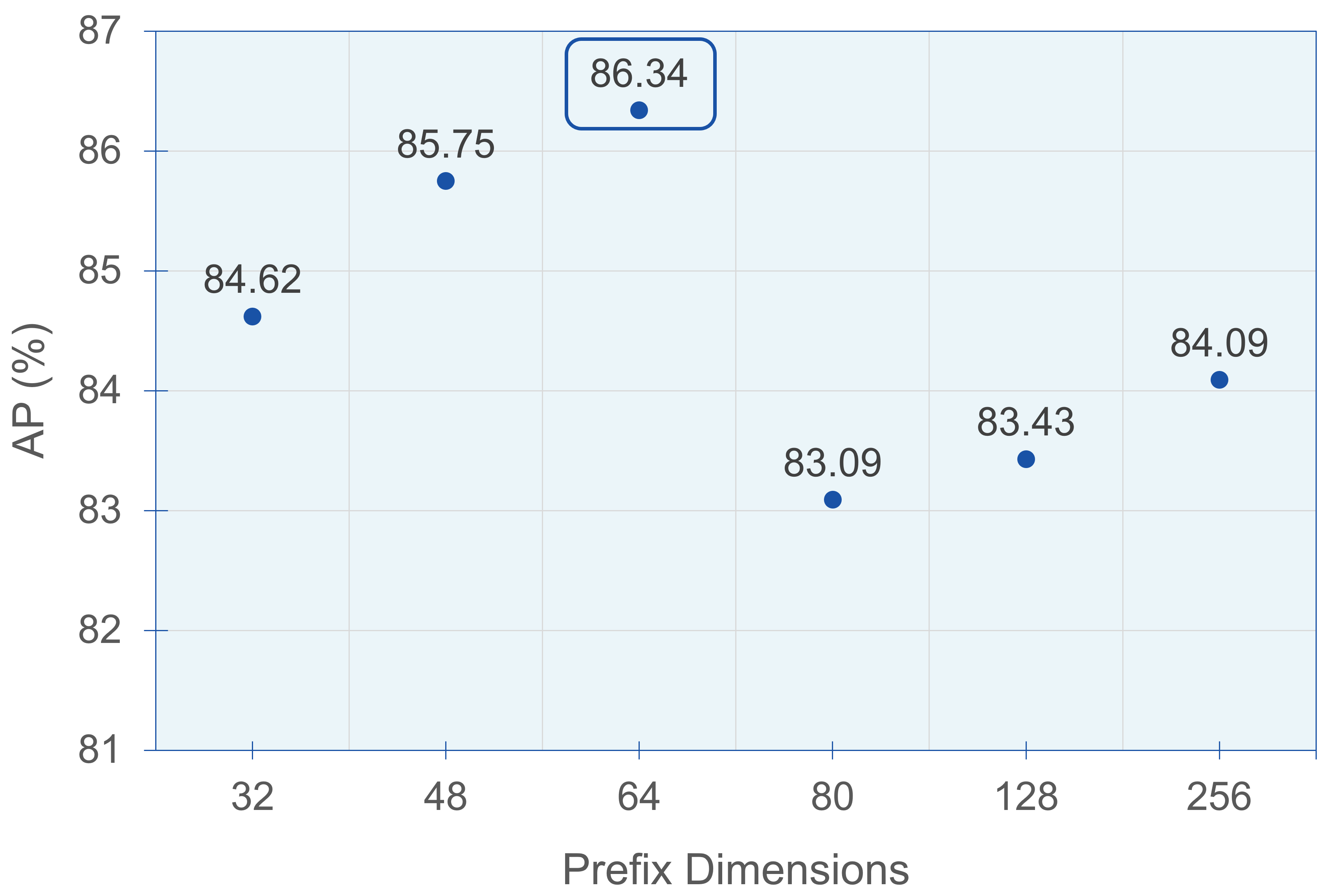}
    \caption{Ablation studies on different setting of prefix dimensions $D_p$ on our proposed CFA module}
    \label{fig:ablation2}
\end{figure}

\begin{figure}
    \centering
    \includegraphics[width=\linewidth]{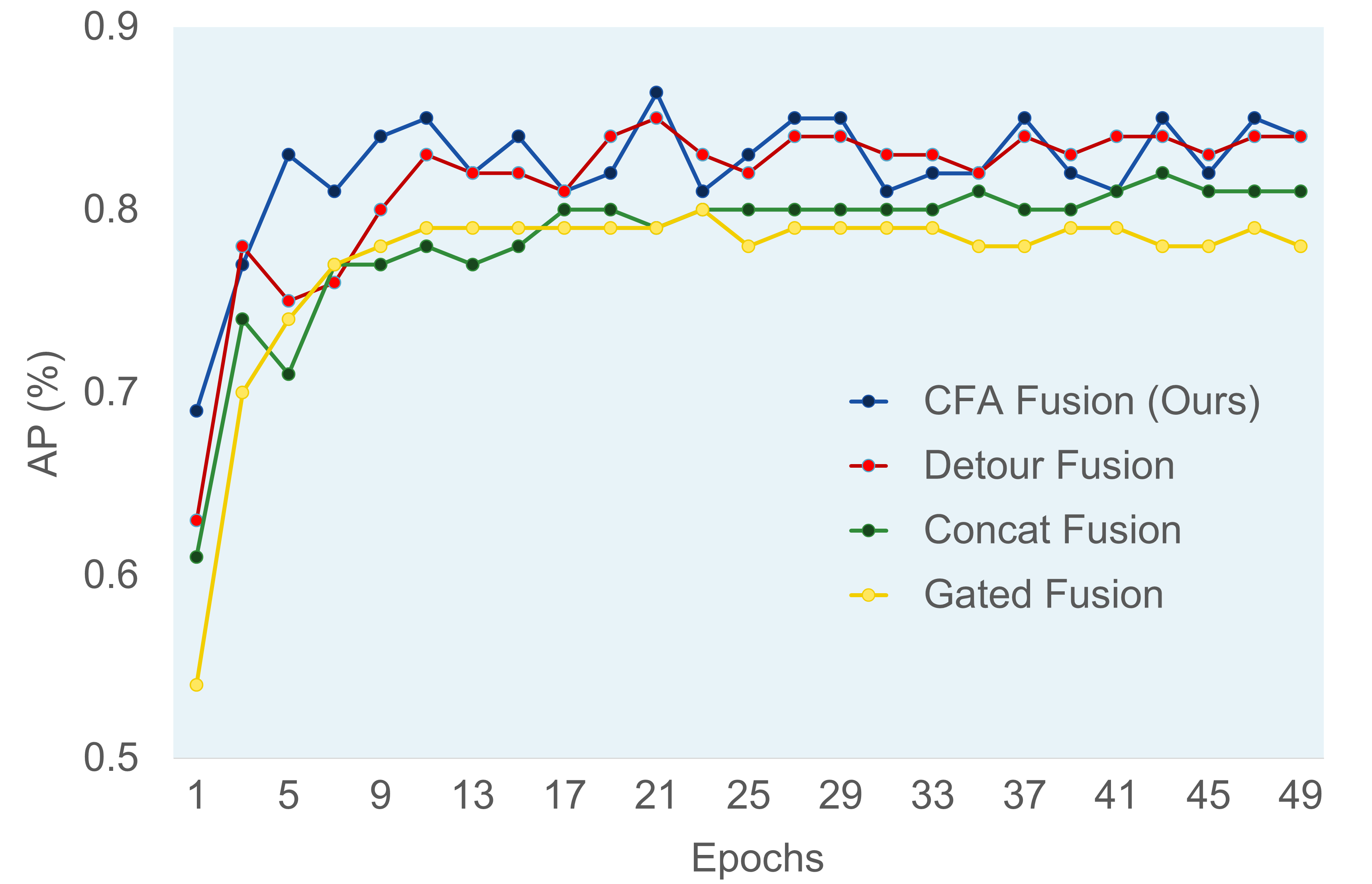}
    \caption{Comparative analysis (Epoch vs AP) with existing fusion methods with our proposed CFA fusion.}
    \label{fig:ablation3}
\end{figure}

In order to ensure a fair comparison, all ablation experiments were conducted using the XD-Violence testing dataset for the violence detection task.
\\
\textbf{Analysis of proposed components:} A series of ablation experiments were conducted to validate the efficacy of the proposed components, i.e., CFA and HLGAtt, within the proposed framework. Few experiments have been performed on the CFA module with and without Prefix-Tuning (\textit{PT}) and Modulation Mechanism (\textit{MM}) modules. The results are outlined in Table \ref{Tab:ablation1}. It is evident that the inclusion of the \textit{PT} and \textit{MM} modules enhances the performance of the CFA module. Furthermore, experiments were carried out using the attention mechanism FHGCN, as proposed in \cite{HyperVD}, to validate the effectiveness of the HLGAtt module. In Table \ref{Tab:ablation1}, we present the results performed using the FHGCN module in Cases 1 - 3, while Cases 4 - 6 show the result obtained from the proposed HLGAtt module. The HLGAtt module demonstrates superior performance compared to the FHGCN \cite{HyperVD} module by a significant margin. For instance, there is a 4.69\% improvement from Case 1 to Case 4, a 4.58\% improvement from Case 2 to Case 5, and a 2.92\% improvement from Case 3 to Case 6. This analysis proves the efficacy of the proposed modules.  
\begin{table}[t]
\centering
\caption{Ablation studies on introduced components i.e., CFA \& HLGAtt of the proposed framework. Here, PT indicates Prefix-Tuning, MM indicate Modulation Mechanism.} \label{Tab:ablation1}
\vspace{-0.5em}
\begin{adjustbox}{max width=0.99\linewidth}
\begin{tabular}{ccccccc}
\hline
\textbf{Cases} & \textbf{CFA w/o PT \& MM}    & \textbf{CFA w/o PT}  & \textbf{CFA}  & \textbf{FHGCN \cite{HyperVD}} & \textbf{HLGAtt}      & \textbf{AP (\%)}    \\ \hline
 \rowcolor{lightmintbg}\rule{-2pt}{10pt} 1             &  \checkmark     &    &    &  \checkmark &   &    81.02          \\
\rule{-2pt}{10pt} 2              &           &  \checkmark  &     &   \checkmark   &    &     81.51 \\
\rowcolor{lightmintbg}\rule{-2pt}{10pt} 3              &           &    & \checkmark    &  \checkmark &   &     83.42     \\
\rule{-2pt}{10pt}  4              &  \checkmark         &    &  &    &   \checkmark          &      85.71     \\
\rowcolor{lightmintbg}\rule{-2pt}{10pt}  5              &          &  \checkmark   &  &    &   \checkmark          &          86.09 \\
 \rule{-2pt}{10pt} 6     &   &  & \checkmark   &   &  \checkmark   & \textbf{86.34} \\ \hline
\end{tabular}
\end{adjustbox}
\end{table}
\\
\noindent\textbf{Analysis on prefix dimension in the Prefix-Tuner module:} 
We also thoroughly analyze the prefix dimension within the Prefix-Tuner module featured in the CFA module. The experiment tested different prefix dimensions including 34, 48, 64, 80, 128, and 256. The corresponding results are illustrated in Figure \ref{fig:ablation2} where one can observe that the optimal performance is achieved as AP of 86.34\%, with a prefix dimension of 64. However, the performance of AP is diminished when the prefix dimension is increased beyond 64. As a result, we set the prefix dimension as 64 in the proposed CFA module.
\\
\noindent\textbf{Analysis of different fusion mechanism:} A series of ablation experiments were conducted to evaluate the efficacy of the proposed CFA fusion mechanism in comparison to alternative fusion methods like Detour fusion \cite{HyperVD}, Concat fusion, and Gated fusion. The impact of the AP measure obtained from these experiments during the training process is illustrated in Figure \ref{fig:ablation3}. It can be seen from this analysis that the proposed CFA fusion module consistently achieves superior performance across all fusion mechanisms.

\section{Conclusion}
This research presents a new WS-VAD framework with a Cross-modal Fusion Adapter (CFA) module and a Hyperbolic Lorentzian Graph Attention (HLGAtt) module to detect anomaly events such as violence and nudity accurately. The CFA module addresses the imbalanced modality information issue and effectively facilitates multi-modal interaction by dynamically selecting the relevant audio features with corresponding visual features. Additionally, the HLGAtt module captures the hierarchical relationships within normal and abnormal representations, thereby improving the accuracy of separating normal and abnormal features. Through extensive experiments and ablation studies, it has been demonstrated that the proposed model outperforms existing violence and nudity detection methods.

{
\small
\bibliographystyle{ieeenat_fullname}
\bibliography{references}
}
\balance

\end{document}